\documentclass{article}
\usepackage{graphics}

\begin{document}

\title{Two-Dimensional Cellular Automata and the Analysis of Correlated Time
Series}

\author{Lu\'\i s~O.~Rigo~Jr.\\
Valmir~C.~Barbosa\thanks{Corresponding author (valmir@cos.ufrj.br).}\\
\\
Universidade Federal do Rio de Janeiro\\
Programa de Engenharia de Sistemas e Computa\c c\~ao, COPPE\\
Caixa Postal 68511\\
21941-972 Rio de Janeiro - RJ, Brazil}

\date{}

\maketitle

\begin{abstract}
Correlated time series are time series that, by virtue of the underlying process
to which they refer, are expected to influence each other strongly. We introduce
a novel approach to handle such time series, one that models their interaction
as a two-dimensional cellular automaton and therefore allows them to be treated
as a single entity. We apply our approach to the problems of filling gaps and
predicting values in rainfall time series. Computational results show that the
new approach compares favorably to Kalman smoothing and filtering.

\end{abstract}

\section{Introduction}\label{intr}

Let $P$ be a set comprising $p$ members, and for each $i\in P$ consider the
sequence $X_i=\langle x_i^1,\ldots,x_i^T\rangle$, where $x_i^t$ is a real
number for $1\le t\le T$. In this paper we consider scenarios in which the
sequences $X_1,\ldots,X_p$ are correlated time series, that is, the $t$ that
provides indices into each sequence is a time parameter, and moreover for
distinct $i$ and $j$ the constituents of $X_i$ cannot be assumed to be
independent of those of $X_j$.

Typical situations in which such a scenario arises are those in which $P$ stands
for a set of points in geographic space (assumed flat, for simplicity) and each
$X_i$ stands for a series of periodic measurements related to some natural
process at point $i$, such as rainfall, temperature, and several others.
Important problems related to the processing of such time series are the
estimation of missing values and also the prediction of values before they are
measured. The former problem is normally posed on the set of full sequences,
that is, after they have each acquired $T$ entries, even though some of these
entries may in fact be tags for missing values, henceforth called gaps. The
latter problem, in turn, can be posed for each point $i$ at all instants
$t=2,\ldots,T$ and requires that $x_i^t$ be predicted after only $t-1$ instants
have elapsed.

Despite the fact that $X_1,\ldots,X_p$ are dependent on one another, the usual
approach to either of the two problems mentioned above is to handle each
sequence separately via some of the several known methods of time-series
completion or prediction, as the case may be \cite{wg94}. According to this
approach, filling a gap in sequence $i$ at time $t$ (that is, estimating the
missing $x_i^t$) results from a function of the non-gap values in $X_i$.
Likewise, predicting $x_i^t$ is achieved as a function of
$x_i^1,\ldots,x_i^{t-1}$. To the best of our knowledge, no approaches have yet
been put forward that allows for the expansion of such dependencies to reflect
the underlying reality that the $p$ sequences are in fact correlated time
series. Clearly, an approach resulting from such an expansion would compute the
missing $x_i^t$ as a function of the non-gap values in all of $X_1,\ldots,X_p$
or yet predict the value of $x_i^t$ as a function of the first $t-1$ values in
all the $p$ sequences.

Our contribution in this paper is to introduce a new approach to the treatment
of correlated time series. Our approach can be used for both gap filling and
value prediction. Qualitatively, it relies on expanding the dependencies alluded
to above so that the inherent correlation between constituents of distinct time
series is taken into account. As its core premise, we postulate the existence
of functions $f_1,\ldots,f_p$ such that, for $1\le i\le p$ and $t>1$,
$x_i^t=f_i(x_1^{t-1},\ldots,x_p^{t-1})$. In essence, this is to say that we do
take inter-series dependencies into account, but do so in a sort of
``memoryless'' framework that lets values corresponding to time $t$ depend on
past values only as far back as $t-1$.

When the points in $P$ are located in some two-dimensional space, as they are in
the examples mentioned above, the postulated functions $f_1,\ldots,f_p$ can be
regarded as the update functions of a hybrid two-dimensional cellular automaton.
We review these automata in Section~\ref{ca}, and from then on they provide
the abstraction to be used in the remainder of the paper. Given the appropriate
cellular automaton, we show in Section~\ref{fillpred} how to use it for filling
gaps in the time series and also for predicting future values. Determining the
cellular automaton, however, requires in essence that we find suitable functions
$f_1,\ldots,f_p$. In Section~\ref{genetic}, we formulate this problem as a
problem of learning from examples and show how to solve it by a genetic
algorithm. We then proceed to a discussion of computational results on rainfall
time series in Section~\ref{results} and close in Section~\ref{concl} with
concluding remarks.

\section{Two-dimensional cellular automata}\label{ca}

Cellular automata are discrete-time abstract devices \cite{i01}. They were
introduced decades ago as models of computation and are currently thought by
many to be the quintessential model for the emergence of complex behavior in
several domains, including various of the fundamental processes of nature
\cite{w02}. The one-dimensional variants of cellular automata have been the ones
to be most widely and deeply studied \cite{w94}. They are for this reason the
best known variants, even though the two-dimensional variant known as the Life
game is highly popular \cite{bcg82}.

In this section we steer our review directly toward the two-dimensional case,
which is the one that, as we will see, relates closely to the time-series
problems we are considering. Normally a two-dimensional cellular automaton is
defined by placing a simple processing element (a cell) at each of the nodes of
a two-dimensional grid. In this brief review we start with a more general cell
placement and let one cell exist for each of the members of $P$, which is
henceforth used to denote the set of cells as well. Each cell $i$ has a
neighborhood within $P$, which is a subset of $P$ that we denote by $N_i$ and
necessarily includes $i$ itself. We let $n_i=\vert N_i\vert$.

Each cell $i$ is a simple automaton whose state at discrete time $t$, for
$t\ge 1$, we denote by $s_i^t$ and let be given by one of the members of a
discrete set $S$ (common to all cells). Starting at the initial states
$s_1^1,\ldots,s_1^p$, the $p$ cells evolve synchronously in time in such a way
that, for $t>1$, $s_i^t$ is a function of every $s_j^{t-1}$ such that
$j\in N_i$. This function is the so-called update rule for cell $i$; if we
denote it by $g_i$ and let $N_i=\{j_1,\ldots,j_{n_i}\}$, then the evolution of
the cell's state is such that
$s_i^t=g_i(s_{j_1}^{t-1},\ldots,s_{j_{n_i}}^{t-1})$.

As we look back on the time series $X_1,\ldots,X_p$ introduced in
Section~\ref{intr}, the correspondence between $X_i$ and the state sequence
$\langle s_i^1,\ldots,s_i^T\rangle$ becomes clear if only we allow each $x_i^t$
occurring in $X_i$ to be approximately represented by the integer giving the
interval into which $x_i^t$ falls in some discretization of its range. This
given, all that the correspondence requires is that we equate each $x_i^t$ with
$s_i^t$, and similarly each $f_i$ with $g_i$, provided $S$ is the set of
integers implied by the underlying discretization .

We then see that, in principle, two-dimensional cellular automata provide a
suitable abstraction of our new approach to handling correlated time series.
The crux of the approach, therefore, is now shifted toward finding appropriate
functions $g_1,\ldots,g_p$. Each $g_i$ is a map leading from $S^{n_i}$ to $S$,
so the number of different possibilities for $g_i$ is $s^{s^{n_i}}$, where
$s=\vert S\vert$. This quantity becomes unthinkably large very quickly as $n_i$
is increased, so finding $g_i$ may very quickly become an impossible task. Also,
it is conceivable that representing $g_i$ once it is determined requires as much
as $O(s^{n_i})$ space, which also grows exponentially with $n_i$.

One common approach to try and curb such explosive growth is to follow Life's
update-rule style and adopt the so-called outer-totalistic update rules
\cite{pw85}. We say that $g_i$ is outer-totalistic if it is a function of
$s_i^{t-1}$ and of the sum of the remaining $n_i-1$ states of the cells in $N_i$
at time $t-1$. What we do in this paper is to partition $N_i\setminus\{i\}$ into
the $q_i\ge 1$ sets $N_i^{(1)},\ldots,N_i^{(q_i)}$, respectively of sizes
$n_i^{(1)},\ldots,n_i^{(q_i)}$, and then to generalize outer-totality as
follows. The update rule $g_i$ is made to depend on $s_i^{t-1}$ and on $q_i$
sums of states at time $t-1$, each computed inside one of the sets
$N_i^{(1)},\ldots,N_i^{(q_i)}$. Once this is done, the number of distinct
possibilities for $g_i$ becomes $s^{n_i^{(1)}\ldots n_i^{(q_i)}s^{1+q_i}}$; the
worst-case space required for representing $g_i$, likewise, becomes
$O(n_i^{(1)}\ldots n_i^{(q_i)}s^{1+q_i})$.

However, it is very important to note that, once an outer-totalistic rule style
is adopted, the correspondence between the set of $p$ time series and the
synchronous evolution of the two-dimensional cellular automaton on $p$ cells has
to be reexamined carefully. The reason is that this correspondence depends
crucially on the set $S$ that summarizes the discretization of the original real
numbers, but adding up members of $S$ retains no meaning in the time-series
setting. We then henceforth assume that, even though $S$ remains an invariant
set, the meaning of each of the $s$ intervals it represents depends on which
quantity is being referred to. If such a quantity is one of the original
time-series elements, then the intervals' meaning is as we have discussed. But
when we refer to a sum of states for outer-totality, then we assume that first
the sum is computed on the original real numbers and only then is discretization
applied. If the number of states to be summed up is $z$, then the range of such
discretization is $z$ times that of the individual numbers.

Adopting our generalized outer-totalistic style substitutes an exponential
dependency on $q_i$ for one on $n_i$. While we expect this to have substantial
impact on the search for $g_i$, representing each of $g_1,\ldots,g_p$ may still
be overly costly. However, being as they are necessarily learned from examples,
each of these functions is in all likelihood only approximately representable,
as the diversity of examples to represent them otherwise is itself exponentially
large with $q_i$ and not very many examples are in general available. Moreover,
whichever examples we have to work with are derived from in-situ measurements,
which are themselves subject to error.

Based on these considerations, we split the determination of the update rule
$g_i$ into two phases. The first phase promotes the learning from examples of
$N_i$ together with the value of $q_i$ and the sets
$N_i^{(1)},\ldots,N_i^{(q_i)}$. We note that the number of possible outcomes for
fixed $n_i$ amounts to what is called the $(n_i-1)$th Bell number \cite{gkp94}.
We postpone discussing this phase until Section~\ref{genetic}.

The second phase is the determination of $g_i$ itself once $N_i$, $q_i$, and the
sets $N_i^{(1)},\ldots,N_i^{(q_i)}$ have been determined. This determination is
dependent upon the specific problem to be solved, as for example the problems of
gap filling and value prediction discussed in Section~\ref{intr}. In any event,
we choose to represent $g_i$ as a table $T_i$ to be computed from the input
corpus available for the problem at hand. Each row in this table corresponds to
a possible input to $g_i$ (but not conversely, since many possible inputs are
likely to be unrepresented in that corpus), that is, to a sequence of $1+q_i$
integers. Also, the table has three columns, denoted as follows at row $r$:
$I_i^r$ is the input itself, $O_i^r$ is an output on that input, and $C_i^r$ is
a count related to the occurrence of the pair $(I_i^r,O_i^r)$ in the input
corpus (it is meant to account for the aforementioned possibility of data
inconsistency).

In what follows, we discuss how to build and use this table in the cases of gap
filling and of value prediction. The neighborhood $N_i$ and the partition of
$N_i\setminus\{i\}$ into $N_i^{(1)},\ldots,N_i^{(q_i)}$ are assumed to be known
and to remain fixed throughout.

\section{Gap filling and value prediction}\label{fillpred}

\subsection{Gap filling}\label{fill}

For gap filling we assume that the sequence $\langle s_i^1,\ldots,s_i^T\rangle$
is known beforehand for every $i\in P$, and also which entries are tags for
gaps. These $p$ sequences can then be regarded as referring to actual data
coming from $T$ successive measurement rounds performed on some underlying
process at the $p$ points.

The first step is to build the tables $T_1,\ldots,T_p$, which is achieved as
follows. For $t=2,\ldots,T$ and $i=1,\ldots,p$, if $s_i^t$ is not a gap then we
check whether the input to cell $i$ corresponding to time $t-1$ depends on none
of the gaps that may exist in $s_1^{t-1},\ldots,s_p^{t-1}$. In the affirmative
case, we compute the $q_i$ sums that the partition dictates. If the resulting
input is already present in $T_i$ with $s_i^t$ as output on some row $r$, then
we simply increment $C_i^r$. If it is not present, then a new row $r$ is added
to $T_i$ with $I_i^r$ given by that input, $O_i^r=s_i^t$, and $C_i^r=1$.

Having completed all $p$ tables, we proceed to filling gaps. We do so by
stepping $t$ upward from $2$ through $T$ and for each value of $t$ examining
each cell $i$ such that $s_i^t$ is a gap. If the input to the cell can be
obtained from $s_1^{t-1},\ldots,s_p^{t-1}$ without involving any gaps, then we
let $r$ be the row of $T_i$ for which $I_i^r$ is closest, by Hamming distance,
to this input. If more than one row exists, then we pick the row $r$ among them
for which $C_i^r$ is greatest. Having selected the appropriate $r$, we fill the
gap by letting $s_i^t=O_i^r$. We then revise all $p$ tables to reflect this new
value and move on to the next value of $t$.

Notice that some gaps may remain unfilled, including at least those for $t=1$.
In Section~\ref{results} we consider a few alternatives to try and tackle this.

\subsection{Value prediction}\label{pred}

For value prediction we start with the $p$ values $s_1^1,\ldots,s_p^1$ that
correspond to $t=1$ and build the tables $T_1,\ldots,T_p$ incrementally as $t$
is stepped upward from $2$ through $T$. We assume that none of
$s_1^1,\ldots,s_p^1$ are gaps, even though this cannot in general be guaranteed
(cf.\ Section~\ref{results} for further alternatives). As in Section~\ref{fill},
we regard these $p$ initial values as coming from measurement data on some
underlying process. We also assume that $p$ new values (possibly including gaps)
become available at each new value of $t$. Our task is to try to predict them
before they become available.

For each new value of $t$, first the $p$ tables are updated in exactly the same
fashion as in Section~\ref{fill}. For this update, each of
$s_1^{t-1},\ldots,s_p^{t-1}$ corresponds to an actual measurement value, unless
$t>2$ and that value is in fact a gap, in which case the value predicted for
instant $t-1$ is used instead. Then we let $i=1,\ldots,p$ and predict that the
upcoming $s_i^t$ will have value $O_i^r$, where $r$ is the row of $T_i$ selected
also as in Section~\ref{fill}. Clearly, the assumed absence of gaps for $t=1$
ensures that a prediction can be made at all subsequent instants for all cells.

\section{Determining a cell's neighborhood and its partition}\label{genetic}

As we noted earlier, the determination of cell $i$'s neighborhood $N_i$ and of
the partition $\{N_i^{(1)},\ldots,N_i^{(q_i)}\}$ is approached as learning from
examples in a training set. The example corpus we use as training set comprises,
for each cell $i$, the sequence $\langle s_i^1,\ldots,s_i^T\rangle$, which may
contain gaps. For each $i$, we start with a set $M_i\subseteq P$ of size $m_i$
such that $i\in M_i$, and a fixed value for $n_i$ such that $n_i\le m_i$. We
then proceed to selecting $N_i$ from the size-$n_i$ subsets of $M_i$ that
include $i$ as a member. Starting at such a superset $M_i$ is a means of
ensuring that the eventual $N_i$ will comply with certain requirements
pertaining to the nature of the time series at hand, as for example those
related to a physical region's geography. We also set an upper bound on the
eventual value of $q_i$; this upper bound is denoted by $u_i$ and is intended to
set limits on the exponential behavior that is inherent to Bell-number growth.

For fixed $N_i$, $q_i$, and $\{N_i^{(1)},\ldots,N_i^{(q_i)}\}$, a score related
to such neighborhood and partition can be computed from the assumed set of
examples as follows. First we build the table $T_i$; this is done as in
Section~\ref{fill}, so examples for which $N_i$ contains gaps are skipped. If
$T_i$ ends up having $L$ distinct inputs among its rows, then the score is the
number in the interval $[0,1]$ given by
\begin{equation}
\varphi(\mathcal{N}_i)=\frac{1}{Z}\sum_{l=1}^Lw_lC_i^l,
\label{fitness}
\end{equation}
where $\mathcal{N}_i$ refers to the triple
$\langle N_i,q_i,\{N_i^{(1)},\ldots,N_i^{(q_i)}\}\rangle$. In (\ref{fitness}),
$C_i^l$ is the greatest of the $C_i^r$'s in $T_i$ that correspond to the $l$th
distinct input, while $w_l$ is an application-related weight, as will be
exemplified in Section~\ref{results}. The $Z$ dividing the summation is needed
to keep the score within $[0,1]$ and is given by the sum of $w_lC_i^r$ for $r$
ranging over the entire $T_i$ and $l$ indicating which of the $L$ distinct
inputs row $r$ corresponds to.

The score $\varphi(\mathcal{N}_i)$ grows with the internal consistency of
$\mathcal{N}_i$ vis-\`a-vis the set of examples. In other words, cell $i$'s
neighborhood and its partition lead to a higher score in proportion to how
consistently occurrences of the same input to $i$ within the examples imply the
same output. We may then view the problem of determining the cell's neighborhood
and its partition as the problem of optimizing $\varphi(\mathcal{N}_i)$ over all
the possibilities for $\mathcal{N}_i$.

Such an optimization problem is of course highly unstructured and also
non-differentiable, so in this paper our approach to solving it is to employ a
genetic algorithm that operates on individuals representing the various
possibilities for $\mathcal{N}_i$ and seeks the one that is fittest according
to the measure of fitness given by $\varphi(\mathcal{N}_i)$. We take each
individual to be a sequence of $m_i-1$ integers, each corresponding to each of
the members of $M_i\setminus\{i\}$ (the potential neighbors of cell $i$, itself
excluded). Of these integers, $m_i-n_i$ are $0$ and indicate the cells that are
not in the set $N_i\setminus\{i\}$ according to this individual. The other
$n_i-1$ integers come from the set $\{1,\ldots,u_i\}$, each indicating the
partition set to which the corresponding cell belongs; the number of distinct
integers occurring amid these $n_i-1$ integers is the value of $q_i$ according
to this individual.

The genetic algorithm we use in our experiments of Section~\ref{results} is one
of the common variants of the generational genetic algorithm \cite{m96}. It goes
through a fixed number of generations, each comprising a fixed number of
individuals, and at the end outputs the fittest individual ever encountered.
Each new generation is obtained from the previous one by first performing an
elitist step whereby a fraction of that generation's fittest individuals is
copied directly to the new one. Then the new generation is filled by individuals
selected from the previous generation after they undergo either crossover (as a
pair) or mutation (individually).

We perform selection randomly in proportion to the individuals' linearly
normalized fitness scores. That is, if $K$ is the fixed size for each
generation, then the $k$th fittest individual, with $1\le k\le K$, is selected
with probability proportional to
\begin{equation}
\Phi-\left(\frac{\Phi-1}{K-1}\right)(k-1),
\label{nfitness}
\end{equation}
where $\Phi$ is a parameter indicating how likely the fittest of the $K$
individuals is to be selected when compared to the least fit one.

The crossover of two individuals employs a random binary mask and yields two
offspring: the first inherits the integers marked $0$ on the first parent by the
mask and those marked $1$ on the second parent; the second offspring inherits
the complementary integers from each parent. Obviously it is possible for an
offspring to have more than $n_i-1$ nonzero integers, in which case it is
corrected by setting randomly chosen nonzero integers to $0$. As for the
mutation of an individual, it is performed on a randomly chosen integer by
mutating it into any member of $\{0,\ldots,u_i\}$ (i.e., the corresponding cell
may be removed from cell $i$'s neighborhood, if it is there to begin with, or be
assigned to any of the possible partition sets).

\section{Computational experiments}\label{results}

\begin{figure}
\centering
\begin{tabular}{c@{\hspace{0.00in}}c}
\scalebox{0.50}{\includegraphics{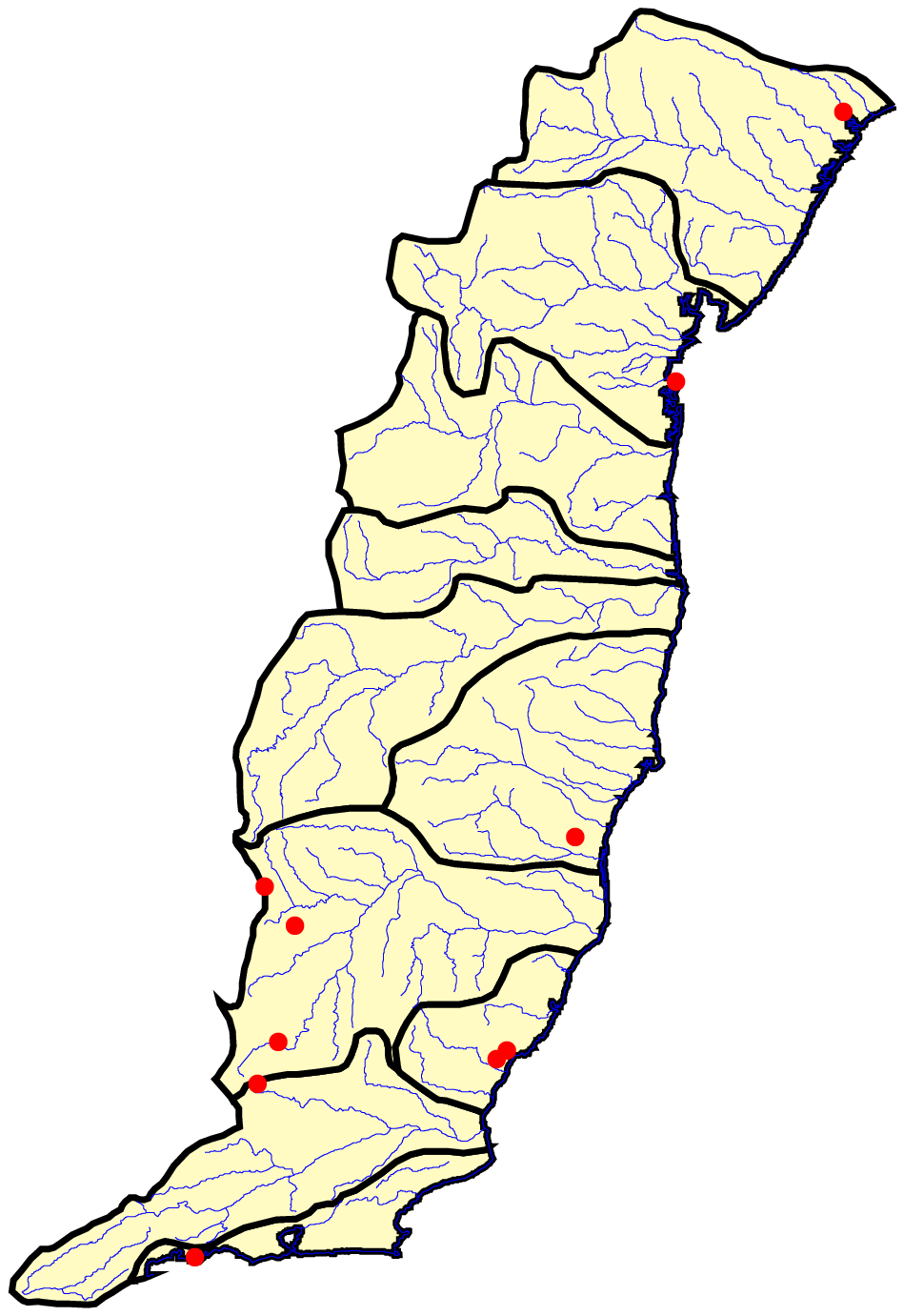}}&
\scalebox{0.50}{\includegraphics{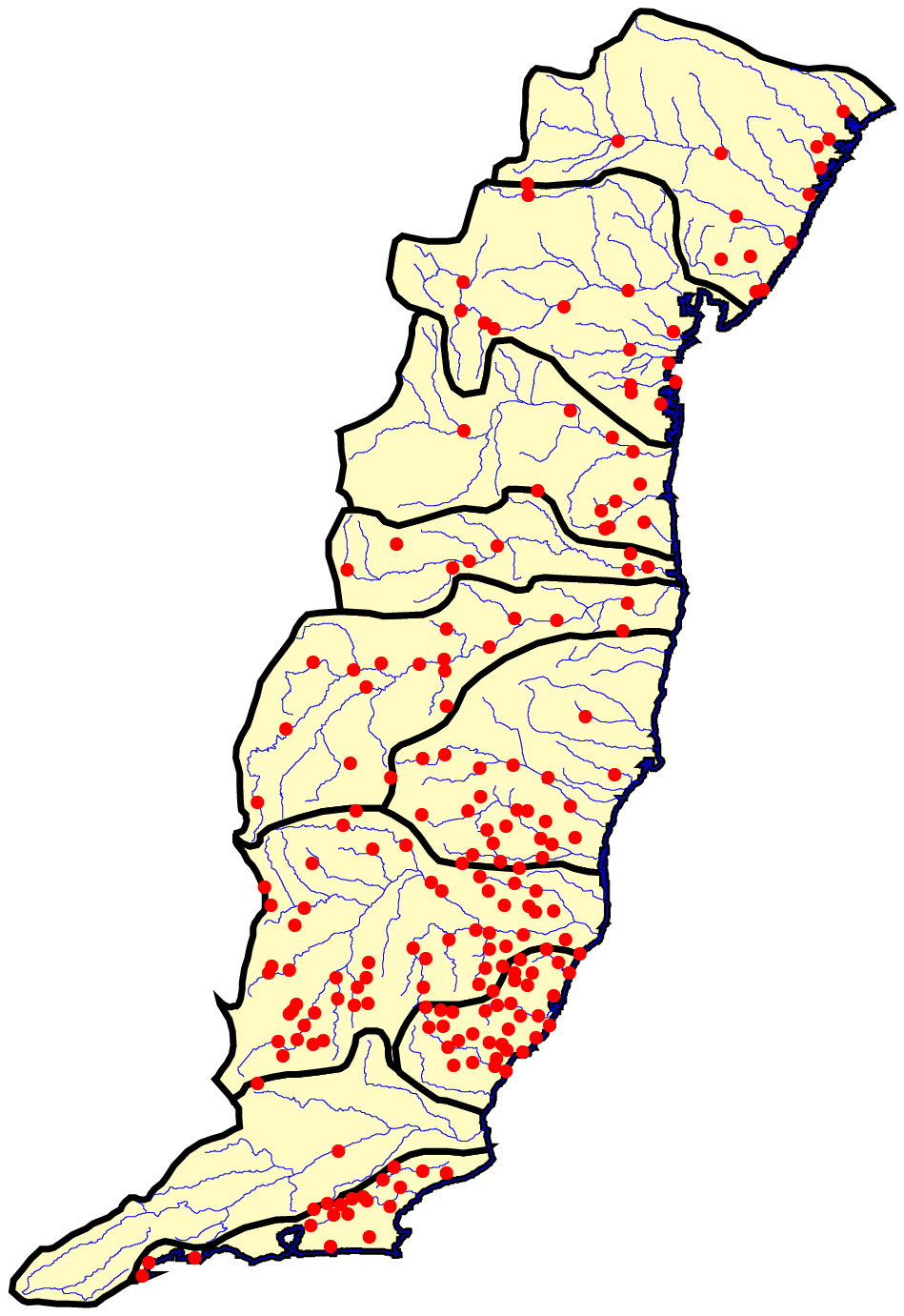}}\\
(a)&
(b)
\end{tabular}
\caption{Brazil's eastern Atlantic basin. Rainfall measurement sites are shown
as filled circles: the $10$ sites in part (a) constitute the set $P$, while the
$207$ sites in part (b) are those out of which neighborhoods may be determined
during the genetic algorithm. Thick lines delimit sub-basins, thin lines
represent watercourses.}
\label{basin}
\end{figure}

The problem domain we have selected for illustrating our approach is that of
rainfall time series. The results we report in this section refer to the eastern
Atlantic basin in Brazil \cite{ana}, shown in the maps of Figure~\ref{basin}.
This region contains $551$ measurement sites, of which $207$ have a gap fraction
of no more than $0.1$. All series refer to daily measurements on the twenty-year
period of 1981--2000, so $T=7305$. All data are available as fixed-point numbers
and we have chosen $S=\{0,\ldots,9\}$ in such a way that $0$ corresponds to the
absence of rainfall and the remaining $9$ values correspond to equally wide
intervals of increasingly large rainfall figures within the appropriate range
(recall from Section~\ref{ca} that this range may refer to individual rainfall
figures or to combined figures for outer-totality).

The training set we use with the genetic algorithm of Section~\ref{genetic} is
selected from the $207$ time series by first choosing $10$ of them to constitute
the set $P$ (cf.\ Figure~\ref{basin}) and then randomly choosing $5\%$ of the
non-gap entries from each of these $10$ series and turning them into gaps. The
entries replaced by these artificially added gaps constitute the test set we use
for evaluating the performance of the overall approach. We note that, even
though $P$ contains only $10$ cells, each cell's potential neighborhood may
include, in principle, any of the $207$ cells. This is in slight disaccord with
our description in preceding sections, but we proceed in this way in order to
avoid an excessively large experiment.

Not all rainfall intervals are represented equally in the training set. In fact,
there is an almost overwhelming predominance of the low-end interval $0$, which
represents the absence of rain precipitation. Such imbalance is known to be
problematic as far as the learning performed by the genetic algorithm is
concerned, so we use the weights appearing in (\ref{fitness}) to compensate.
Specifically, if we recall that $L$ is the number of distinct inputs
occurring in $T_i$, and furthermore that $C_i^l$ is the greatest count occurring
in $T_i$ for the $l$th input, then we let $w_l$ be obtained from the following
linear normalization of those counts. Let us first say, for simplicity's sake,
that $C_i^1$ is the least of the $L$ counts, and so on through $C_i^L$ being the
greatest of them. Then we let
\begin{equation}
w_l=W-\left(\frac{W-1}{L-1}\right)(l-1),
\label{weight}
\end{equation}
where $W$ is a parameter indicating the ratio of the largest weight to the
smallest. Clearly, weights set in this manner are such that the rarest rainfall
interval receives the largest weight ($W$), and so on through the most common
interval receiving weight $1$.

\begin{figure}
\centering
\scalebox{0.50}{\includegraphics{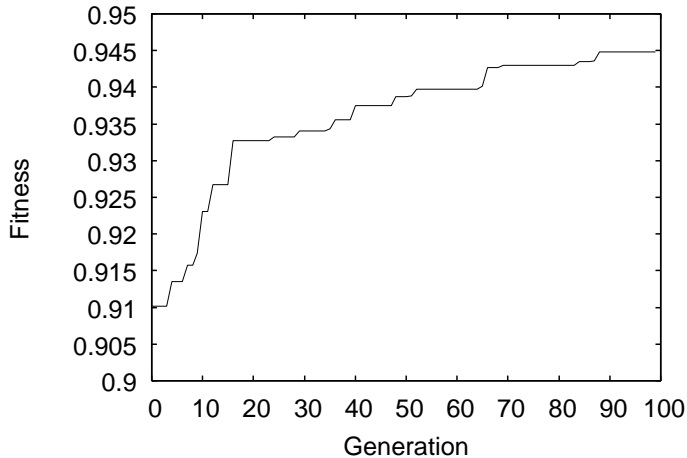}}
\caption{Evolution of the fitness of (\ref{fitness}) for one of the cells in
$P$.}
\label{evolution}
\end{figure}

Each of our runs of the genetic algorithm on cell $i\in P$ produces $100$
generations, each containing $1000$ individuals. We also let $M_i$ comprise
nearest neighbors by Euclidean distance in such a way that $m_i=30$. In
addition, each run uses $n_i=20$, $u_i=15$, an elite rate of $0.02$, and a
probability of $0.5$ for deciding between crossover and mutation. We also let
$\Phi=40$ and $W=5$ in (\ref{nfitness}) and (\ref{weight}), respectively. A
typical evolution of the fitness given by (\ref{fitness}) is shown in
Figure~\ref{evolution}.
 
Our results for filling or predicting the artificially inserted gaps that
constitute the test set are shown in Tables~\ref{caf}--\ref{kfp}. Each table
shows an overall hit ratio (the fraction of gaps that are correctly filled or
predicted) and also a hit ratio for each of the possible intervals in $S$ (the
fraction of gaps within each of the intervals that are correctly filled or
predicted). These latter hit ratios are only shown for intervals $0$--$4$,
since none of the other intervals is represented in the test set.
Tables~\ref{caf} and \ref{ksf} refer to gap filling, respectively by cellular
automata and by Kalman smoothing \cite{wb}. Tables~\ref{cap} and \ref{kfp} refer
to value prediction, respectively by cellular automata and Kalman filtering
\cite{wb}. Results of the two Kalman procedures come from the code implemented
in \cite{m04} with the number of iterations parameter set to $10$. Note that
Kalman filtering and smoothing, unlike our cellular automata, operate in a
manner that is confined to each time series individually. They also operate
directly on the original fixed-point input values; for the sake of comparing
their results to those obtained by the cellular automata, their outputs are
first cast into the same intervals represented by the set $S$.

As we remarked in Section~\ref{fillpred}, the two basic procedures described in
that section for gap filling and value prediction may be unsuccessful due to the
fact that cellular-automaton update rules are represented in the succinct,
approximate format of the tables $T_i$. When this is the case, the results given
in Tables~\ref{caf} and \ref{cap} already reflect the following attempts at
improvement. First, the $5$ fittest individuals output by the genetic algorithm
are used in succession, as opposed to using the one fittest individual only,
until no gap is left unfilled or unpredicted. If still not enough, then all $5$
individuals are once again considered, now with cell neighborhoods diminished by
the removal of exactly one cell (the one to have the least impact on the
unweighted version of (\ref{fitness})). This process of neighborhood diminishing
proceeds while feasible.

In all four tables we use the recourse of highlighting strictly best figures
with a bold typeface. These refer to comparing Tables~\ref{caf} and \ref{ksf},
and also Tables~\ref{cap} and \ref{kfp}. Comparing all results in this manner
reveals that our cellular automata tend to perform better in overall terms than
the corresponding Kalman procedures. As we examine the rainfall intervals
individually, the cellular automata are seen to be still ahead, but now the
Kalman procedures are best performers in several occasions as well. Notice,
interestingly, that the entries in which Kalman smoothing or filtering has a
ratio superior to the corresponding cellular automaton refer invariably to
intervals $0$ or $1$, which are by far the most common rainfall intervals. The
cellular automata, by contrast, are on occasion successful in interval $2$, too.

\section{Concluding remarks}\label{concl}

We have introduced two-dimensional cellular automata as an abstraction for
handling multiple correlated time series. The method that results from this
abstraction is based on learning succinct, approximate versions of the cellular
automata's update rules from examples. This seems to be the first attempt at
handling correlated time series concomitantly so that problems such as gap
filling and value prediction can take into account the series' interrelatedness.
The resulting cellular automaton, if successful, can then be regarded as an
approximate model of the physical reality underlying the observed data in the
time series.

We have provided computational results on the problems of gap filling and value
prediction in the domain of rainfall time series. These results compare
favorably to the preeminent procedures of Kalman smoothing and filtering, the
former applied on gap filling, the latter on value prediction.

\begin{table}[p]
\centering
\caption{Results for gap filling by cellular automata.}
\label{caf}
    \begin{tabular}{ccccccc}
      \hline & & \multicolumn{5}{c}{Hit ratio per interval} \\
      \cline{3-7} Cell & Overall hit ratio & 0 & 1 & 2 & 3 & 4 \\ \hline
      1 & \textbf{0.717} & \textbf{0.736} & 0.705 & 0.000 & 0.000 \\
      2 & \textbf{0.748} & 0.688 & \textbf{0.805} & 0.000 &  & \\
      3 & \textbf{0.718} & \textbf{0.824} & \textbf{0.537} & 0.000 &  & \\
      4 & \textbf{0.721} & 0.840 & \textbf{0.531} & \textbf{0.167} &  & \\
      5 & \textbf{0.777} & \textbf{0.865} & 0.453 & 0.000 & 0.000 & \\
      6 & \textbf{0.725} & 0.803 & \textbf{0.600} & \textbf{0.167} &  & \\
      7 & \textbf{0.750} & \textbf{0.870} & 0.246 & 0.000 &  & \\
      8 & \textbf{0.729} & 0.822 & \textbf{0.570} & \textbf{0.182} &  & \\
      9 & \textbf{0.691} & 0.792 & \textbf{0.536} & 0.000 & 0.000 & 0.000 \\
      10 & \textbf{0.623} & \textbf{0.682} & 0.523 & 0.000 &  & \\
      \hline
    \end{tabular}

\end{table}

\begin{table}[p]
\centering
\caption{Results for gap filling by Kalman smoothing.}
\label{ksf}
    \begin{tabular}{ccccccc}
      \hline & & \multicolumn{5}{c}{Hit ratio per interval} \\
      \cline{3-7} Cell & Overall hit ratio & 0 & 1 & 2 & 3 & 4 \\ \hline
      1 & 0.344 & 0.000 & \textbf{1.000} & 0.000 & 0.000 \\
      2 & 0.397 & \textbf{0.917} & 0.005 & 0.000 &  & \\
      3 & 0.485 & 0.550 & 0.374 & 0.000 &  & \\
      4 & 0.633 & \textbf{1.000} & 0.000 & 0.000 &  & \\
      5 & 0.216 & 0.015 & \textbf{1.000} & 0.000 & 0.000 & \\
      6 & 0.654 & \textbf{1.000} & 0.000 & 0.000 &  & \\
      7 & 0.191 & 0.016 & \textbf{1.000} & 0.000 &  & \\
      8 & 0.658 & \textbf{0.972} & 0.000 & 0.000 &  & \\
      9 & 0.654 & \textbf{1.000} & 0.000 & 0.000 & 0.000 & 0.000 \\
      10 & 0.477 & 0.415 & \textbf{0.633} & 0.000 &  & \\
      \hline
    \end{tabular}

\end{table}

\clearpage

\begin{table}[p]
\centering
\caption{Results for value prediction by cellular automata.}
\label{cap}
    \begin{tabular}{ccccccc}
      \hline & & \multicolumn{5}{c}{Hit ratio per interval} \\
      \cline{3-7} Cell & Overall hit ratio & 0 & 1 & 2 & 3 & 4 \\ \hline
      1 & \textbf{0.708} & \textbf{0.729} & \textbf{0.685} & 0.250 & 0.000 \\
      2 & \textbf{0.723} & 0.643 & \textbf{0.795} & 0.000 &  & \\
      3 & \textbf{0.699} & \textbf{0.782} & 0.553 & \textbf{0.250} &  & \\
      4 & \textbf{0.712} & 0.797 & \textbf{0.594} & 0.000 &  & \\
      5 & \textbf{0.768} & \textbf{0.844} & 0.488 & 0.000 & 0.000 & \\
      6 & \textbf{0.720} & \textbf{0.815} & 0.567 & 0.000 &  & \\
      7 & \textbf{0.732} & \textbf{0.844} & \textbf{0.261} & 0.000 &  & \\
      8 & \textbf{0.723} & 0.814 & \textbf{0.570} & \textbf{0.182} &  & \\
      9 & \textbf{0.711} & 0.802 & \textbf{0.571} & \textbf{0.125} & 0.000 & 0.000 \\
      10 & 0.617 & 0.708 & \textbf{0.450} & 0.000 &  & \\
      \hline
    \end{tabular}

\end{table}

\begin{table}[p]
\centering
\caption{Results for value prediction by Kalman filtering.}
\label{kfp}
    \begin{tabular}{ccccccc}
      \hline & & \multicolumn{5}{c}{Hit ratio per interval} \\
      \cline{3-7} Cell & Overall hit ratio & 0 & 1 & 2 & 3 & 4 \\ \hline
      1 & 0.507 & 0.484 & 0.562 & 0.250 & 0.000 \\
      2 & 0.430 & \textbf{1.000} & 0.000 & 0.000 &  & \\
      3 & 0.337 & 0.000 & \textbf{1.000} & 0.000 &  & \\
      4 & 0.633 & \textbf{1.000} & 0.000 & 0.000 &  & \\
      5 & 0.223 & 0.030 & \textbf{0.977} & 0.000 & 0.000 & \\
      6 & 0.497 & 0.475 & 0.567 & 0.000 &  & \\
      7 & 0.446 & 0.533 & 0.072 & 0.000 &  & \\
      8 & 0.677 & \textbf{1.000} & 0.000 & 0.000 &  & \\
      9 & 0.654 & \textbf{1.000} & 0.000 & 0.000 & 0.000 & 0.000 \\
      10 & \textbf{0.674} & 1.000 & 0.000 & 0.000 &  & \\
      \hline
    \end{tabular}

\end{table}

\clearpage

\subsection*{Acknowledgments}

The authors acknowledge partial support from CNPq, CAPES, and a FAPERJ BBP
grant.

\bibliography{catseries}

\begin{thebibliography}{10}

\bibitem{bcg82}
E.~R. Berlekamp, J.~H. Conway, and R.~K. Guy.
\newblock {\em Winning Ways for Your Mathematical Plays}, volume~2.
\newblock Academic Press, London, UK, 1982.

\bibitem{gkp94}
R.~L. Graham, D.~E. Knuth, and O.~Patashnik.
\newblock {\em Concrete Mathematics}.
\newblock Addison-Wesley, Boston, MA, second edition, 1994.

\bibitem{ana}
Hidro{W}eb: {H}ydrological {I}nformation {S}ystem.
\newblock http://{\linebreak[0]}hidroweb.{\linebreak[0]}%
  ana.{\linebreak[0]}gov.{\linebreak[0]}br.
\newblock In Portuguese.

\bibitem{i01}
A.~Ilachinski.
\newblock {\em Cellular Automata}.
\newblock World Scientific, Singapore, 2001.

\bibitem{m96}
M.~Mitchell.
\newblock {\em An Introduction to Genetic Algorithms}.
\newblock The MIT Press, Cambridge, MA, 1996.

\bibitem{m04}
K.~Murphy.
\newblock Kalman filter toolbox for {M}atlab.
\newblock http://{\linebreak[0]}www.{\linebreak[0]}cs.%
  {\linebreak[0]}ubc.{\linebreak[0]}ca/{\linebreak[0]}%
  {\kern-.5pt\lower3pt\hbox{\char'176}\kern.5pt}murphyk/{\linebreak[0]}%
  Software/{\linebreak[0]}Kalman/{\linebreak[0]}kalman.html.

\bibitem{pw85}
N.~H. Packard and S.~Wolfram.
\newblock Two-dimensional cellular automata.
\newblock {\em Journal of Statistical Physics}, 38:901--946, 1985.

\bibitem{wg94}
A.~S. Weigend and N.~A. Gershenfeld, editors.
\newblock {\em Time Series Prediction}.
\newblock Perseus Books, Reading, MA, 1994.

\bibitem{wb}
G.~Welch and G.~Bishop.
\newblock The {K}alman filter.
\newblock http://{\linebreak[0]}www.{\linebreak[0]}%
  cs.{\linebreak[0]}unc.{\linebreak[0]}edu/{\linebreak[0]}%
  {\kern-.5pt\lower3pt\hbox{\char'176}\kern.5pt}welch/{\linebreak[0]}kalman.

\bibitem{w94}
S.~Wolfram.
\newblock {\em Cellular Automata and Complexity}.
\newblock Addison-Wesley, Reading, MA, 1994.

\bibitem{w02}
S.~Wolfram.
\newblock {\em A New Kind of Science}.
\newblock Wolfram Media, Champaign, IL, 2002.

\end{thebibliography}
\bibliographystyle{plain}

\end{document}